\documentclass[sigplan,10pt]{acmart}
\usepackage{graphicx}
\usepackage{subcaption}
\usepackage{caption}
\usepackage{multirow}
\usepackage{pifont}
\usepackage{enumitem}

\setcopyright{none}
\renewcommand\footnotetextcopyrightpermission[1]{}

\title{PulseInfer: I/O-Centric Sparse KV Cache Offloading for Efficient Long-Context LLM Decoding}

\author{Qiuyang Zhang}
\authornote{Both authors contributed equally to this research.}
\affiliation{%
  \institution{Huazhong University of Science and Technology}
  \city{Wuhan}
  \country{China}
}
\email{qyzhang@hust.edu.cn}

\author{Kai Zhou}
\authornotemark[1]
\affiliation{%
  \institution{Huazhong University of Science and Technology}
  \city{Wuhan}
  \country{China}
}
\email{kezzy@hust.edu.cn}

\author{Kai Lu}
\authornote{Corresponding Author: Kai Lu (kailu@hust.edu.cn)}
\affiliation{%
  \institution{Huazhong University of Science and Technology}
  \city{Wuhan}
  \country{China}
}
\email{kailu@hust.edu.cn}

\author{Haocheng Lu}
\affiliation{%
  \institution{Huazhong University of Science and Technology}
  \city{Wuhan}
  \country{China}
}
\email{haocheng_lu@hust.edu.cn}

\author{Jian Zhou}
\affiliation{%
  \institution{Huazhong University of Science and Technology}
  \city{Wuhan}
  \country{China}
}
\email{jianzhou@hust.edu.cn}

\author{Yuanpeng Su}
\affiliation{%
  \institution{UCloud Technology Co., Ltd.}
  \city{Shanghai}
  \country{China}
}
\email{chris.su@ucloud.cn}

\author{Kun Bao}
\affiliation{%
  \institution{UCloud Technology Co., Ltd.}
  \city{Shanghai}
  \country{China}
}
\email{kun.bao@ucloud.cn}

\author{Jiguang Wan}
\affiliation{%
  \institution{Huazhong University of Science and Technology}
  \city{Wuhan}
  \country{China}
}
\email{jgwan@hust.edu.cn}

\author{Fei Wu}
\affiliation{%
  \institution{Huazhong University of Science and Technology}
  \city{Wuhan}
  \country{China}
}
\email{wufei@mail.hust.edu.cn}

\begin{document}

\pagestyle{plain}

\begin{abstract}

Long-context LLM serving is increasingly bottlenecked by decode, where large KV caches limit batch size and underutilize GPUs. Sparse KV cache offloading expands effective capacity by storing most historical KV blocks in CPU DRAM and recalling only selected blocks on demand. However, we find that existing offloading systems shift the bottleneck to CPU-GPU recall I/O: recall volume varies widely across layers, decode steps and requests, while headwise sparse selection fragments recalls into many small PCIe transfers.

This paper presents PulseInfer, an I/O-centric sparse KV cache offloading system. PulseInfer hides variable recall latency with interruptible layer-wise scheduling, adapts offloading decisions with IO-Adaptive Offloading Admission, and coalesces fragmented transfers using SoloHead sparse selection and a gather-scatter I/O engine. Implemented on SGLang, PulseInfer improves decode throughput by up to 4.7× over SGLang and 2.6× over the best existing offloading baseline, while reducing TPOT by up to 76\% and preserving near-lossless accuracy.
\end{abstract}

\maketitle

\section{Introduction}

Large language model (LLM) serving has become a dominant workload in modern data centers~\cite{aegaeon,serverlessllm, kvcache-in-the-wild,cachedattention,llm-serving-survey}, driven by the widespread adoption of chatbots~\cite{gpt}, code assistants~\cite{claude-code,cursor,codex}, and personal AI agents~\cite{openclaw,claude-cowork}.
Meanwhile, two trends are reshaping inference workloads.
On the input side, context windows have expanded from thousands of tokens to hundreds of thousands or even millions, enabling long-document understanding, retrieval-augmented generation, and multi-turn agentic histories~\cite{rag}.
On the output side, reasoning models~\cite{deepseek-r1,gpt-o1} and inference-time scaling~\cite{cot, test-time-survey} make generation increasingly long, with models often producing thousands of chain-of-thought tokens before the final answer.
Together, longer inputs and outputs shift inference cost toward the \emph{decode} phase, which executes once per generated token over an ever-growing context and can account for over 90\% of end-to-end inference time.

High-throughput decoding relies on large batch sizes to amortize per-token overheads and fully utilize GPU compute.
However, long-context requests require large KV caches whose memory footprint grows linearly with context length, leaving GPU memory space for only a few concurrent requests.
This batch-size constraint makes long-context decoding \emph{memory-capacity-bound}, severely degrading serving throughput.

Sparse KV cache offloading is a promising approach to this problem~\cite{arkvale,infinigen,pqcache}.
Instead of keeping the entire KV cache in GPU memory, it offloads most historical KV blocks to CPU DRAM while retaining only digests and hot blocks on the GPU.
During decode, a sparse attention selector identifies important historical blocks for the current query, and the system recalls missing KV blocks from CPU DRAM to GPU memory before attention computation.
By offloading temporarily unimportant tokens to CPU, this design extends the effective KV capacity beyond GPU memory.

However, KV cache offloading shifts the bottleneck from GPU memory capacity to CPU--GPU recall I/O.
For every layer in every decode step, the system must transfer missing KV blocks from host DRAM to GPU memory over PCIe, whose effective bandwidth is one to two orders of magnitude lower than HBM, especially for small transfers.
In synchronous recall systems such as ArkVale~\cite{arkvale} and PQCache~\cite{pqcache}, the GPU stalls until the required blocks arrive, often driving GPU utilization to around 35\%.
Even InfiniGen~\cite{infinigen}, which prefetches KV entries one layer ahead, achieves only around 45\% GPU utilization under realistic serving workloads, because a static pipeline cannot fully hide variable recall latency.
As a result, recall-I/O stalls can offset the capacity benefit of offloading: under realistic workloads, existing KV cache offloading systems often fail to improve throughput and may even underperform non-offloading baselines.

We find that this poor utilization stems from two fundamental I/O pathologies in existing sparse offloading systems.
\textbf{First, the volume of recall I/O is inherently dynamic.}
The number of recalled tokens varies across decode steps, layers, requests, and batches because sparse selection is query-dependent and workload composition changes over time.
For a single request, recall I/O follows a highly right-skewed, long-tailed distribution: most layers and steps require moderate recall, while a non-negligible fraction incurs much larger transfers.
This distribution also varies significantly across layers, with median recall load differing by up to 5$\times$.
At the serving level, shifting request-length distributions further change the number of offloaded sequences in each batch, causing aggregate recall I/O to fluctuate over time.
Such variability makes recall latency difficult to hide with a fixed pipeline.

\textbf{Second, the granularity of recall I/O is too small.}
Existing sparse attention mechanisms select important tokens independently for each KV head, producing many small, scattered transfers rather than a few large contiguous ones.
This headwise granularity is algorithmically natural but systemically inefficient: it drastically reduces effective PCIe bandwidth, reaching only around 6 GB/s on PCIe 5.0$\times$16.
It also requires each head to keep its own digest metadata resident in GPU memory.
Together, these pathologies turn recall I/O into a critical-path stall, causing KV cache offloading to underutilize expensive GPUs and reduce system throughput.

To address these issues, we propose \textbf{PulseInfer}, an I/O-centric KV cache offloading system for high-throughput long-context LLM decoding.
PulseInfer rethinks KV cache offloading from an I/O perspective: recall I/O is not a secondary concern, but the central bottleneck that must be \textbf{scheduled}, \textbf{controlled}, and \textbf{coalesced}.
Accordingly, PulseInfer decomposes the problem into three goals: hiding variable recall latency, adapting offloading decisions to workload dynamics, and improving the efficiency of each recall transfer.

To hide variable recall latency, PulseInfer introduces \textbf{interruptible layer-wise scheduling}.
Instead of relying on a fixed one-layer-ahead pipeline, PulseInfer partitions requests into normal batches, whose KV caches remain GPU-resident, and offload batches, whose historical KV caches reside in CPU memory, and interleaves their execution at \emph{layer} granularity.
Inspired by the OS interrupt model~\cite{ostep}, PulseInfer treats recall I/O as an asynchronous event rather than a blocking operation.
When an offload batch issues recall requests, GPU execution immediately switches to a normal batch.
The normal batch continues execution while checking for I/O-completion interrupts at layer boundaries, and switches back to the offload batch once recall completes.
This interruptible execution hides variable recall latency behind useful computation while preserving transformer decoding semantics.

To adapt to workload dynamics, PulseInfer introduces \textbf{IO-Adaptive Offloading Admission (IOAA)}.
The effectiveness of interruptible scheduling depends on the balance between normal and offload batches: the normal batch must provide enough computation to absorb the recall I/O generated by the offload batch.
However, as request-length distributions drift over time, any static offloading threshold becomes suboptimal.
Offloading too aggressively creates more recall I/O than the normal batch can hide, while offloading too conservatively leaves GPU memory pressure unresolved.
IOAA formulates batch admission as a throughput-maximization problem.
It builds a quantitative throughput model parameterized by the \textit{overlap depth}, i.e., the number of normal-batch layers needed to cover one round of recall I/O, and searches for the best normal/offload assignment at runtime using a pruned binary-tree search.
Together, IRQ and IOAA allow PulseInfer to hide recall latency effectively under shifting workload mixes, thereby addressing dynamic recall volume.

To improve recall I/O efficiency, PulseInfer coalesces recall I/O at both the algorithm and engine levels.
At the algorithm level, we leverage the observation that attention heads specialize into retrieval heads and streaming heads.
Based on this, we propose \textbf{SoloHead sparse selection}, which uses Middle Attention Mass (MAM) to identify the strongest retrieval head in each layer and lets that head select a shared set of top-$k$ blocks for all KV heads.
SoloHead replaces independent headwise selection with a single layer-level decision, converting fine-grained per-head transfers into coarser per-layer transfers and eliminating per-head digest storage.
By avoiding noisy selections from streaming heads, SoloHead can even improve accuracy over headwise selection across a range of LLM benchmarks.
At the engine level, a \textbf{gather-scatter transfer} engine gathers scattered KV blocks into a pinned staging buffer and transfers them through a single large DMA operation, improving effective PCIe bandwidth by over 2× compared with per-block asynchronous copies.
Together, these mechanisms convert fragmented per-head recall I/O into coarse-grained, bandwidth-efficient transfers.

We implement PulseInfer on SGLang~\cite{sglang} and evaluate it against state-of-the-art baselines on three frontier models ranging from 14B to 230B parameters~\cite{qwen3,minimax-m2.5}, using both synthetic workloads and real industrial traces~\cite{mooncake, servegen}.
PulseInfer improves decode throughput by up to 4.7× over SGLang and 2.6× over the best existing offloading system, sustains \textasciitilde 95\% GPU utilization, and reduces TPOT by up to \textbf{76\%} compared with existing offloading systems, while achieving near-lossless accuracy relative to full attention.

In summary, this paper makes the following contributions:

\begin{itemize}[leftmargin=*]
  \item We identify recall I/O as the key bottleneck of sparse KV cache offloading and characterize its two root pathologies: dynamic recall volume and fragmented transfers.
  \item We propose interruptible layer-wise scheduling (IRQ), an OS-inspired execution model that adaptively hides variable recall latency behind normal-batch computation.
  \item We propose IOAA, a throughput-model-driven policy that dynamically partitions requests between normal and offload batches, and SoloHead with a gather-scatter engine to coalesce fragmented recalls into efficient transfers.
  \item We implement PulseInfer on SGLang, achieving up to 2.6× higher throughput than state-of-the-art offloading systems with near-lossless accuracy.
\end{itemize}

\section{Background}

\subsection{LLM Inference and Sparse Attention}\label{sec:background-llm}

LLM inference consists of a compute-intensive prefill phase and a memory-bound decode phase~\cite{transformer}. During decode, each generated token attends to all previous tokens through the KV cache, whose size grows linearly with sequence length and batch size. Long-context and reasoning workloads therefore make decode increasingly memory-capacity-bound: large KV caches restrict the batch size that can fit in GPU memory, limiting serving throughput. Modern serving systems increasingly adopt prefill-decode (PD) disaggre\/gation~\cite{splitwise,distserve}, where prefill and decode are executed on separate GPU instances to independently optimize compute-intensive and memory-intensive workloads. This paper focuses on the decode instance in such a PD-disaggregated serving setup.

Sparse attention reduces the decode memory traffic by selecting only a subset of historical tokens or blocks for attention~\cite{streamingllm,quest,h2o,nsa,sparse-frontier}. For example, Quest~\cite{quest} partitions the KV cache into fixed-size blocks, maintains compact per-head digests, and lets each KV head independently select its top-$k$ blocks for the current query. While this query-aware selection reduces attention computation, it still assumes that the full KV cache remains GPU-resident.

\subsection{Sparse KV Cache Offloading}\label{sec:background-offloading}

\begin{figure}[tb]
  \centering
  \includegraphics[width=\linewidth]{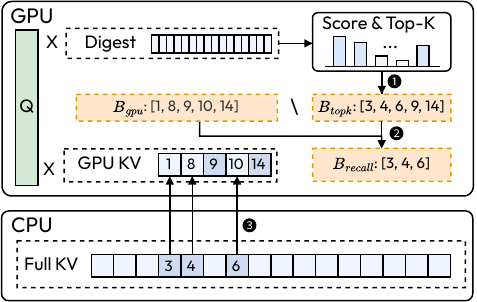}
  \caption{Illustration of sparse KV cache offloading.}
  \label{fig:background-offloading}
\end{figure}

Although sparse attention reduces the amount of memory access during attention computation, it still requires the full KV cache to reside in GPU memory. With the rise of agentic applications~\cite{multi-swe-bench}, inference requests increasingly bundle environmental information, user memory, project files, and multiple rounds of conversation, easily pushing sequence lengths to tens or even hundreds of thousands of tokens and causing the KV cache to become enormous~\cite{react,toolformer,agent-mem,servegen}. For example, a single 128K-token sequence in the Qwen-3 14B model requires 20 GB of KV cache. Because each decoding step performs very little computation, large batch sizes are needed to achieve high overall throughput; however, the substantial KV-cache footprint limits the number of concurrent requests that can fit in GPU memory, making the decode phase memory-capacity-bound.

\begin{figure}[tb]
    \centering
    \begin{minipage}[t]{0.48\columnwidth}
        \centering
        \includegraphics[width=\textwidth]{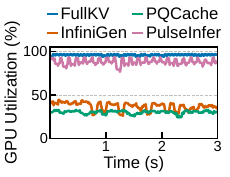}
        \captionof{figure}{GPU utilization across different methods.}
        \label{fig:moti-gpu-util}
    \end{minipage}
    \hfill
    \begin{minipage}[t]{0.48\columnwidth}
        \centering
        \includegraphics[width=\linewidth]{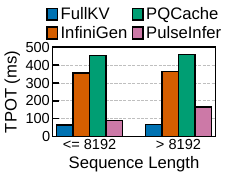}
        \captionof{figure}{TPOT for short and long requests.}
        \label{fig:moti-tpot}
    \end{minipage}
\end{figure}

Sparse KV cache offloading~\cite{arkvale,pqcache,infinigen} addresses this bottleneck by storing the full KV cache in CPU DRAM while keeping only compact digests and frequently accessed hot blocks on the GPU. Building on the sparse-attention workflow, offloading systems add a recall stage. As shown in Figure~\ref{fig:background-offloading}, after sparse selection (\ding{202}) identifies the top-k blocks $B_{topk}$ for the current query, the system computes the set difference between $B_{topk}$ and the GPU-resident hot blocks $B_{gpu}$ (\ding{203}), obtaining the blocks that must be recalled ($B_{recall}$). These missing blocks are then transferred from DRAM to the GPU (\ding{204}).

\begin{figure*}[t]
    \centering
    \begin{minipage}[t]{0.32\textwidth}
        \centering
        \includegraphics[width=\linewidth]{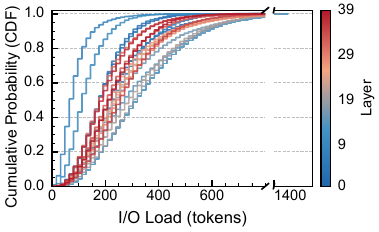}
        \captionof{figure}{Per-layer CDF of recalled tokens on Qwen-3 14B.}
        \label{fig:moti-io-cdf}
    \end{minipage}
    \hfill
    \begin{minipage}[t]{0.32\textwidth}
        \centering
        \includegraphics[width=\textwidth]{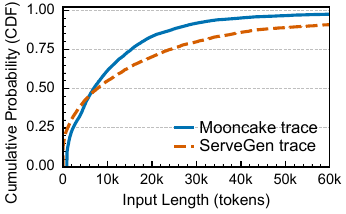}
        \captionof{figure}{Input-length CDFs of realistic serving workloads.}
        \label{fig:moti-workload-cdf}
    \end{minipage}
    \hfill
    \begin{minipage}[t]{0.32\textwidth}
        \centering
        \includegraphics[width=\linewidth]{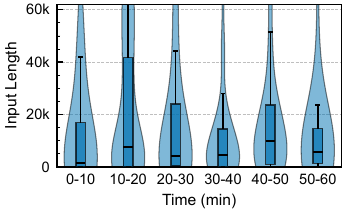}
        \captionof{figure}{Input-length distributions vary over time.}
        \label{fig:moti-workload-shift}
    \end{minipage}
\end{figure*}

For example, ArkVale builds on Quest's~\cite{quest} sparse-attention mechanism (\S\ref{sec:background-llm}) to offload the KV cache. During attention computation, it uses synchronous recall I/O to load the required tokens.
PQCache~\cite{pqcache} formulates KV selection as approximate retrieval and uses product quantization to search offloaded cache.
InfiniGen~\cite{infinigen} further improves KV-cache offloading with one-layer-ahead prefetching, trying to hide I/O latency behind the computation of the preceding layer. To enable this prefetching, InfiniGen~\cite{infinigen} approximates the input to the next layer using the current layer's hidden state, derives the next layer's attention query from this approximation, and predicts the next layer's top-k important blocks with the estimated query to initiate prefetching.

\section{Motivation}

\subsection{Offloading Shifts the Bottleneck to Recall I/O}\label{sec:motivation1}

Sparse KV cache offloading promises to break the GPU memory capacity barrier by storing most of the KV cache on CPU DRAM and recalling only a small subset of important blocks on each decode step. However, this recall is not free~\cite{kvpr}.

Figure~\ref{fig:moti-gpu-util} shows the GPU utilization of two state-of-the-art offloading systems, PQCache~\cite{pqcache} and InfiniGen~\cite{infinigen}, compared to a full-attention baseline. Despite enabling larger batch sizes, both systems suffer from substantial GPU idle time. This idle time stems directly from the GPU waiting for KV cache data to be transferred from CPU memory, highlighting the severity of the recall I/O bottleneck. In effect, offloading shifts the primary bottleneck from GPU memory capacity to recall I/O latency.

Existing systems exacerbate this bottleneck by coupling all requests within a batch. In PQCache and InfiniGen, when any request in the batch triggers recall I/O, the \emph{entire} batch waits. This includes short requests of only a few thousand tokens that require little to no offloaded data and could otherwise complete quickly.
Figure~\ref{fig:moti-tpot} quantifies this head-of-line blocking.
Specifically, we group requests by sequence length to compare their TPOT (time-per-output-token) under mixed-batch offloading against isolated, non-offloaded full attention execution.
Compared with a no-offloading baseline, short requests of fewer than 8K tokens experience a 5× increase in TPOT when co-scheduled with long requests. These requests do not trigger recall I/O themselves, yet they are forced to stall while waiting for recall transfers initiated by longer requests in the same batch.

\subsection{Dynamic Recall I/O Defeats Fixed Prefetching}\label{sec:motivation2}

InfiniGen attempts to hide recall I/O through one-layer-ahead prefetching, using the computation of the current layer to cover the I/O needed by the next layer. As shown in Figure~\ref{fig:moti-gpu-util}, however, this fixed overlap window fails to sustain high GPU utilization. The reason is that recall I/O is dynamic: its volume changes across layers, steps, and requests, making a static one-layer pipeline mismatched to the workload.

Figure~\ref{fig:moti-io-cdf} shows the CDF of recalled tokens across layers under a 2K token budget on Qwen-3 14B. For each layer, the recall I/O load is far from fixed; instead, it follows a highly right-skewed, long-tailed distribution. This is because the attention query changes at every decode step. Although the selected important tokens overlap substantially across steps, the non-overlapping portion varies and is difficult to predict. In addition, the recall-I/O distribution differs significantly across layers, with the median number of recalled tokens varying by up to 5×. This layer-level variation arises from different attention patterns: some layers focus on more stable contextual information, while others attend to more dynamic, query-dependent tokens.
This variability makes it difficult for a fixed prefetch pipeline to consistently hide recall I/O.

\subsection{Workload Drift Breaks Static Offloading Policies}\label{sec:motivation3}

LLM serving workloads introduce another layer of dynamism. First, realistic workloads contain requests with widely different lengths: short conversations of only a few hundred tokens coexist with coding agent requests spanning hundreds of thousands of tokens. Figure~\ref{fig:moti-workload-cdf} shows the request-length CDFs of two publicly available industrial traces, Mooncake~\cite{mooncake} and ServeGen~\cite{servegen}. Across these traces, request lengths routinely span more than three orders of magnitude.

Second, the request-length distribution also changes over time. Figure~\ref{fig:moti-workload-shift} shows the input-length distribution within a one-hour window of the ServeGen trace. Both the median and the spread shift noticeably across windows, indicating that the ratio of short to long requests does not remain stable during serving.

These workload dynamics further amplify recall-I/O variability. To demonstrate this, we evaluate existing offloading systems under two distinct configurations: a synthetic workload with a uniform 32K-token input length, and a real-world workload replayed from the ServeGen trace.
As shown in Figure~\ref{fig:moti-tps}, while InfiniGen and PQCache outperform the full-attention baseline under the uniform synthetic workload, their performance degrades substantially under the real industrial trace, even falling below the baseline.
This contrast underscores that static offloading strategies become inherently fragile when the request mix and recall-I/O demand vary over time.

\subsection{Headwise Selection Fragments Recall I/O}\label{sec:motivation4}

\begin{figure}[tb]
    \centering
    \begin{minipage}[t]{0.48\columnwidth}
        \centering
        \includegraphics[width=\textwidth]{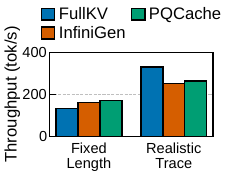}
        \captionof{figure}{Decode throughput under synthetic and real workloads.}
        \label{fig:moti-tps}
    \end{minipage}
    \hfill
    \begin{minipage}[t]{0.48\columnwidth}
        \centering
        \includegraphics[width=\textwidth]{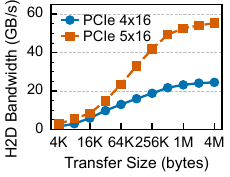}
        \captionof{figure}{Host-to-device PCIe bandwidth under different transfer granularities.}
        \label{fig:moti-pcie}
    \end{minipage}
\end{figure}

Existing sparse-attention methods typically perform headwise selection, where each KV head independently selects its own top-$k$ important tokens or blocks. While algorithmically natural, this design is systemically inefficient for KV offloading. Because different heads often select different blocks, recall I/O is broken into many small, scattered transfers, sometimes as small as a single block for one layer and one KV head. Under a typical configuration with a block size of 16 tokens, a head dimension of 128, and bfloat16 precision, the KV cache for such a transfer is only 8 KB. Such fine-grained transfers are poorly matched to CPU–GPU data movement: as shown in Figure~\ref{fig:moti-pcie}, despite the 64 GB/s theoretical peak bandwidth of PCIe 5.0×16, 8 KB transfers achieve only about 6 GB/s effective bandwidth.

Beyond transfer inefficiency, headwise selection also weakens the memory benefit of offloading. It requires per-block digests to be stored separately for each KV head in GPU memory, and with small block sizes these digests grow into a noticeable fraction of the GPU memory footprint, reducing the savings that offloading would otherwise provide.

\section{Design and Implementation}

\begin{figure}[t]
  \centering
  \includegraphics[width=.8\linewidth]{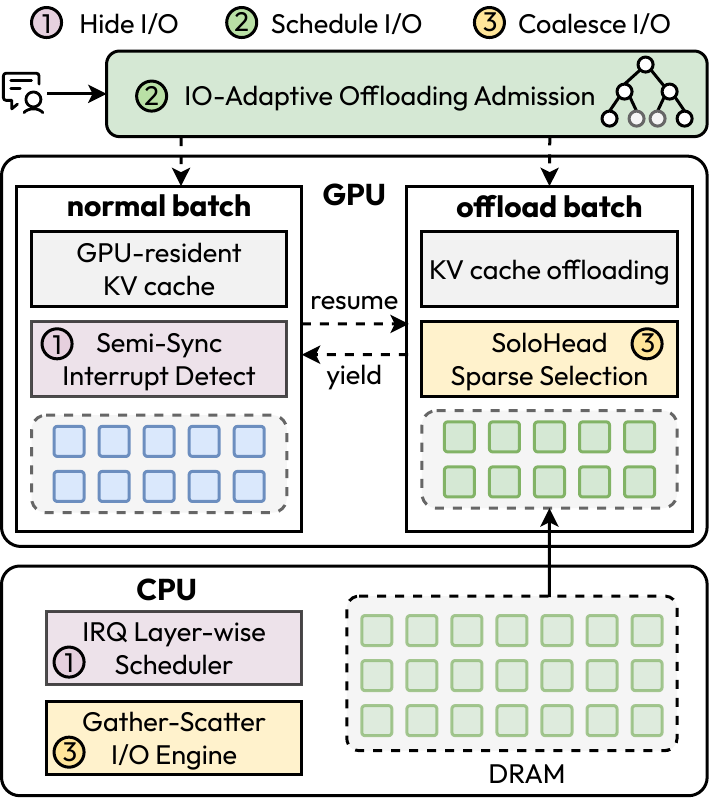}
  \caption{Overall architecture of PulseInfer.}
  \label{fig:arch}
\end{figure}

\subsection{Design Overview}

We propose \textbf{PulseInfer}, an I/O-centric sparse KV cache offloading system that treats recall I/O as a first-class schedulable event. As illustrated in Figure~\ref{fig:arch}, PulseInfer decomposes the I/O bottleneck into three goals---\textbf{hide}, \textbf{control}, and \textbf{coalesce}---each addressed by a dedicated mechanism.

\textbf{Hide (\S\ref{sec:irq}):} PulseInfer partitions requests into a \emph{normal batch} and an \emph{offload batch}, and interleaves them at layer granularity following an OS-style interrupt model, adaptively hiding variable recall I/O behind useful computation.

\textbf{Control (\S\ref{sec:ioaa}):} \emph{IO-Adaptive Offloading Admission (IOAA)} dynamically searches for the normal/offload batch split that maximizes expected throughput under shifting workloads and GPU memory constraints.

\textbf{Coalesce (\S\ref{sec:coalesce}):} \emph{SoloHead} uses a single retrieval head per layer to select a shared top-$k$ block set for all KV heads, and a \emph{gather-scatter transfer engine} coalesces scattered blocks into a single large DMA transfer, jointly improving PCIe bandwidth utilization and reducing digest memory.

\subsection{Hide: Interruptible Layer-wise Scheduling}\label{sec:irq}

As established in \S\ref{sec:motivation1}–\S\ref{sec:motivation2}, recall I/O is the dominant cause of GPU underutilization in KV cache offloading. The fundamental challenge is that recall I/O volume is inherently dynamic: it varies across layers, decode steps, requests, and serving workloads. A fixed prefetch pipeline, such as InfiniGen's one-layer-ahead strategy, cannot reliably absorb this variability.

The problem of hiding variable I/O latency behind useful computation is not unique to LLM serving, it is a well-studied problem in operating systems. When a CPU process issues an I/O request, the OS does not spin-wait until the request completes. Instead, it blocks the waiting process, schedules another ready process, and resumes the original process only after an I/O completion interrupt. This interrupt-driven execution hides unpredictable I/O latency without requiring the OS to predict the duration of each I/O operation.

Inspired by this, we propose \textbf{interruptible layer-wise scheduling (IRQ)} for LLM decoding. The key analogy is straightforward: a CPU process corresponds to an inference batch.
To realize this interrupt-driven model in LLM serving, IRQ classifies decode requests into two types: \textbf{normal requests}, which are relatively short requests that keep their KV cache on the GPU and incur no recall I/O, and \textbf{offload requests}, which are relatively long requests that store their full KV cache in CPU memory while retaining a fixed-size set of hot blocks on the GPU, consisting of sink tokens, local tokens, and hot middle tokens.
This split follows directly from the observation in \S\ref{sec:motivation1}: short requests often require no offloading, yet in existing single-batch systems they are forced to stall while waiting for recall I/O triggered by co-scheduled long requests.
By batching normal and offload requests separately, IRQ overlaps recall I/O with normal-batch computation and avoids exposing short requests to unnecessary I/O stalls.

\begin{figure}[t]
  \centering
  \includegraphics[width=\linewidth]{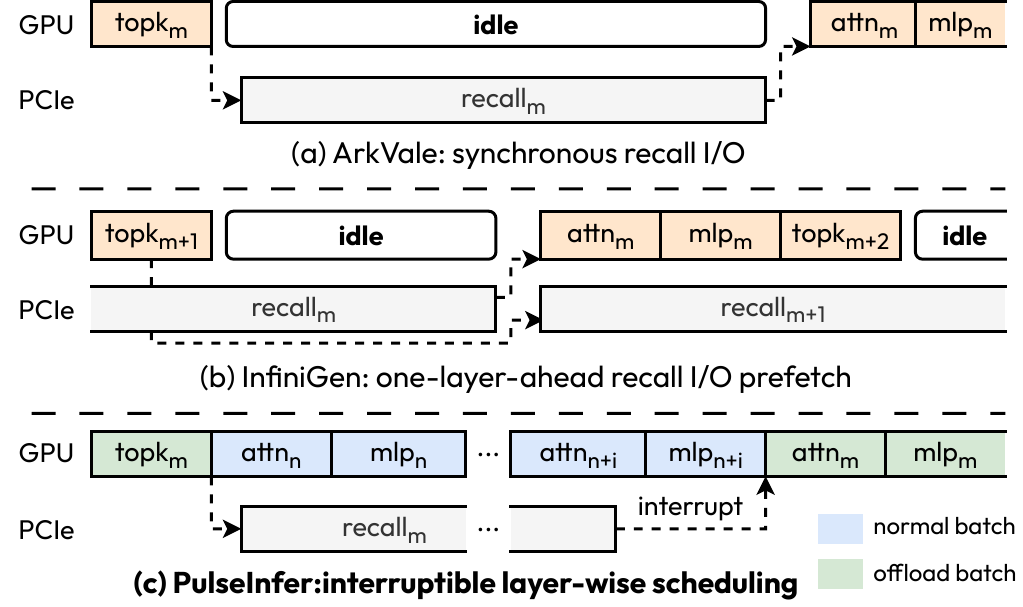}
  \caption{Execution pipelines of different methods.}
  \label{fig:pipeline}
\end{figure}

\subsubsection{Layer-wise Scheduling}\label{sec:layer-wise}

The layer-wise nature of recall I/O motivates a layer-wise scheduler: instead of switching only at token boundaries, IRQ allows execution to switch at transformer-layer boundaries.
Figure~\ref{fig:pipeline} illustrates the resulting execution flow.
The offload batch and the normal batch advance through transformer layers independently. When the offload batch reaches layer $m$, it first performs sparse selection to identify the top-$k$ important blocks and determines which of them are missing from GPU memory. It then submits data-transfer requests to recall the missing KV blocks from CPU memory.
After issuing these transfers, the offload batch yields the GPU, and the scheduler immediately switches to the normal batch, which resumes execution from its current layer $n$.

At each subsequent layer boundary of the normal batch, the scheduler checks whether the offload batch's recall I/O has completed. If the I/O is still in flight, the normal batch proceeds to the next layer and repeats the check.
Once completion is detected, the scheduler switches back to the offload batch. The offload batch then performs attention and FFN computation for layer $m$ using the recalled KV blocks. After finishing this layer, it proceeds to layer $m+1$, performs sparse selection, triggers the next round of recall I/O, and yields the GPU again. The normal batch later resumes from the layer where it was interrupted.

This design has two important properties. First, the \textbf{overlap depth}, defined as the number of normal-batch layers executed during one recall-I/O wait, is not fixed. Instead, it is determined by the actual I/O completion time. This adaptive overlap contrasts with fixed prefetch pipelines, whose overlap window cannot adjust to varying recall loads. Second, the normal batch is isolated from offload-batch I/O. As a result, short requests no longer suffer head-of-line blocking caused by recall transfers that they do not need.

\subsubsection{Semi-Synchronous Interrupt Detection}\label{sec:semi-sync}

\begin{figure}[t]
  \centering
  \includegraphics[width=\linewidth]{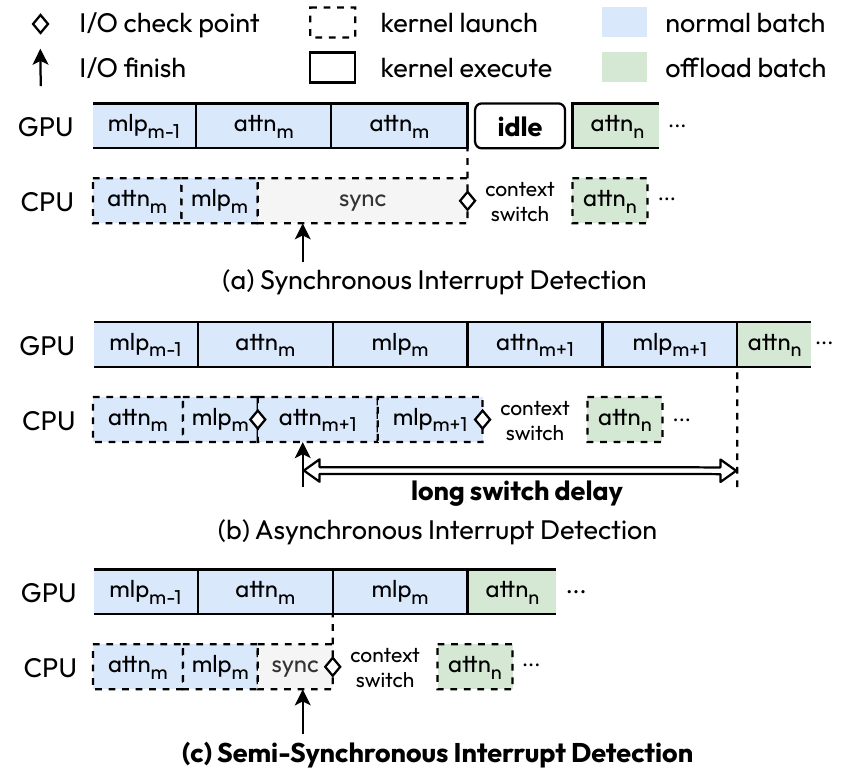}
  \caption{Comparison of interrupt detection methods.}
  \label{fig:semi-sync}
\end{figure}

The interrupt detection mechanism at layer boundaries faces a tension between two goals: minimizing GPU idle time during context switches, and minimizing the delay from I/O completion to batch resumption.
These two goals compete because of the asynchronous nature of GPU execution~\cite{cuda-guide}. GPU tasks are submitted by the CPU into a command queue (e.g., a CUDA stream), and sustaining high GPU utilization requires that this queue never run empty.

A synchronous approach (Figure~\ref{fig:semi-sync}(a)) would require CPU–GPU synchronization at every layer boundary to check I/O completion before deciding the next task. While this keeps the switch timely, the frequent synchronization drains the GPU pipeline and hurts utilization. An asynchronous alternative (Figure~\ref{fig:semi-sync}(b)) avoids synchronization entirely: the CPU keeps pushing normal-batch tasks and polls I/O status between enqueues. This preserves high utilization but delays the context switch---by the time I/O completion is detected, several normal-batch layers may already be queued, forcing the offload batch to wait.

IRQ adopts a \textbf{semi-synchronous interrupt detection} strategy (Figure~\ref{fig:semi-sync}(c)). After submitting the attention and FFN kernels of a normal-batch layer, the CPU waits only for the attention kernel to complete, rather than synchronizing on the entire layer. When this synchronization returns, the GPU has finished attention but still has the FFN kernel queued or running. The scheduler uses this interval to check whether the offload batch's recall I/O has completed and decides which task to enqueue next. By making the scheduling decision while the GPU continues executing FFN, this semi-synchronous design preserves high GPU utilization while bounding the delay of offload-batch resumption.

\subsection{Control: I/O-Adaptive Offloading Admission}\label{sec:ioaa}

\begin{figure*}[t]
    \centering
    \begin{minipage}[t]{0.25\textwidth}
        \centering
        \includegraphics[width=\textwidth]{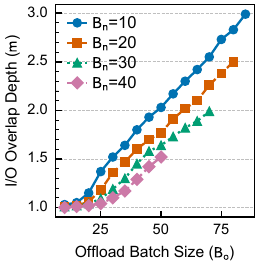}
        \captionof{figure}{Overlap depth of Qwen-3 14B.}
        \label{fig:overlap-depth}
    \end{minipage}
    \hfill
    \begin{minipage}[t]{0.35\textwidth}
        \centering
        \includegraphics[width=\linewidth]{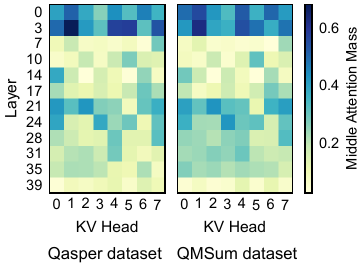}
        \caption{Middle Attention Mass (MAM) across KV heads and layers on Qwen-3 14B.}
        \label{fig:mam-heatmap}
    \end{minipage}
    \hfill
    \begin{minipage}[t]{0.37\textwidth}
        \centering
        \includegraphics[width=\linewidth]{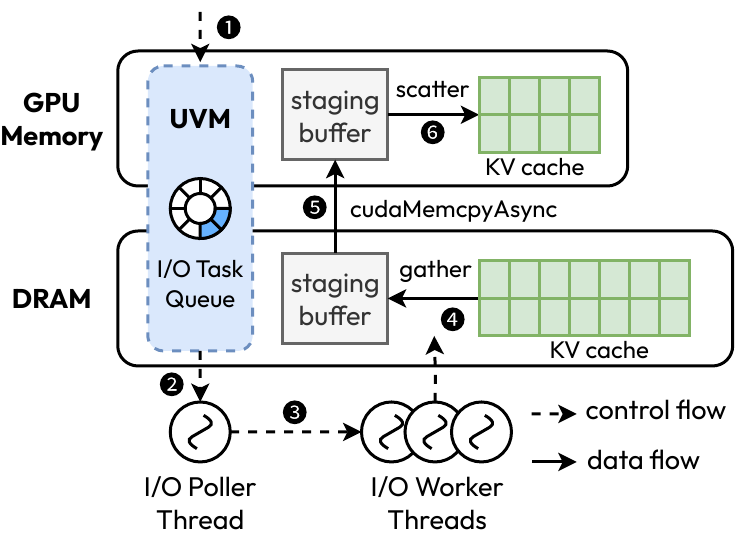}
        \caption{Illustration of gather-scatter transfer (\S\ref{sec:gather-scatter}) and zero-sync I/O submission (\S\ref{sec:impl}).}
        \label{fig:gather-scatter}
    \end{minipage}
\end{figure*}

IRQ scheduling relies on having sufficient normal-batch computation to hide offload-batch recall I/O. However, realistic serving workloads do not provide a stable request mix. As shown in \S\ref{sec:motivation3}, request-length distributions drift over time, causing the ratio of normal to offload candidates to vary continuously. A static length threshold can thus become suboptimal: it may offload too many requests and generate more recall I/O than the normal-batch computation can absorb, or offload too few requests and leave GPU memory pressure unresolved. This motivates a dynamic offloading scheduler that decides which requests to offload based on the current recall-I/O load and GPU memory pressure.

To determine the optimal scheduling policy, we first examine how throughput depends on the sizes of normal and offload batches. During decoding, each request generates one token per step, and step latency varies only mildly with batch size because sparse attention shifts the primary bottleneck away from compute and memory bandwidth. Consequently, system throughput is roughly proportional to the effective batch size. Under IRQ, one offload-batch layer is overlapped with $m$ normal-batch layers, where $m$ denotes the overlap depth. Throughput is therefore approximately proportional to the weighted average of the two batch sizes:
\begin{equation}
\label{eq:throughput}
\text{Throughput} \propto \frac{m}{m+1}\cdot B_n + \frac{1}{m+1}\cdot B_o,
\end{equation}
where $B_n$ and $B_o$ are the normal-batch and offload-batch sizes, respectively.
Since $m \ge 1$, the normal batch consistently receives a larger weight. Holding $m$ fixed, increasing $B_n$ therefore brings a larger throughput gain than increasing $B_o$.

However, the optimal choice is not simply to maximize $B_n$. Increasing $B_o$ can free GPU memory and allow the system to admit a larger total batch size. This can increase either $B_n$ or $B_o$, improving overall throughput. At the same time, a larger offload batch introduces more recall I/O, which drives up the overlap depth $m$. As $m$ grows, throughput becomes increasingly dominated by $B_n$, diminishing the throughput benefit of additional offload requests.

To adapt to the dynamics of workload and balance the tradeoff between offloading and non-offloading decisions, we propose \textbf{IO-Adaptive Offloading Admission (IOAA)}. The core idea is to simulate different admission decisions for the waiting requests using the quantitative throughput model (Eq.~\ref{eq:throughput}), and select the assignment that yields the highest expected throughput.

\subsubsection{Binary-Tree Search with Early Stopping}\label{sec:binary-search}

For a queue of $n$ waiting requests, admission is a combinatorial assignment problem with up to $2^n$ normal/offload configurations. IOAA models this process as a binary search tree, where each level decides the placement of one request, with the two outgoing branches corresponding to normal admission and offloading, and each root-to-leaf path represents a complete assignment. IOAA evaluates Eq.~\ref{eq:throughput} at each feasible leaf and selects the assignment with the highest throughput.

To avoid exhaustive search, IOAA applies two pruning rules. First, branches that exceed the remaining GPU memory budget are pruned immediately. Second, IOAA enforces a length-based monotonicity rule: once a longer request is admitted to the normal batch, shorter requests are not explored for offloading. This matches the threshold structure of the problem and removes unnatural assignments. Together, these rules make the search lightweight: with 30 waiting requests, IOAA visits only 3.16 nodes on average.

\subsubsection{Overlap Depth Profiling}\label{sec:overlap-depth}

The throughput model above requires the overlap depth $m$ for each candidate assignment. By definition, $m = \lceil t_{io}/t_{layer} \rceil$, where $t_{io}$ is the recall I/O latency and $t_{layer}$ is the time to execute one layer of the normal batch. The recall latency $t_{io}$ mainly depends on the number of recalled tokens, which grows with the offload-batch size $B_o$. In contrast, $t_{layer}$ is relatively stable and changes mildly with the normal-batch size $B_n$. Therefore, for a given model and hardware platform, the expected overlap depth can be estimated as a function of $(B_o, B_n)$.

IOAA obtains this function through offline profiling. Figure~\ref{fig:overlap-depth} shows the profiled overlap depth for Qwen-3 14B on an NVIDIA RTX PRO 6000. Since different layers have different recall-load distributions, $m$ is not necessarily an integer; we use the average overlap depth across layers. When $B_o$ is small, recall I/O is short enough to be hidden by one normal-batch layer, so $m$ remains close to 1. As $B_o$ increases, the recall volume grows and $m$ increases nearly linearly. For the same $B_o$, larger $B_n$ slightly increases $t_{layer}$, so the required overlap depth decreases modestly.

Based on this smooth relationship, IOAA profiles $m$ only on a discrete grid of $(B_o, B_n)$ values. At runtime, when evaluating a candidate assignment with a specific $B_o$ and $B_n$, IOAA estimates its overlap depth by interpolating from the nearest profiled points.

\subsection{Coalesce: SoloHead and Gather-Scatter Transfer}\label{sec:coalesce}

\subsubsection{SoloHead Sparse Selection}\label{sec:solohead}

As discussed in \S\ref{sec:motivation4}, existing systems perform sparse selection independently for each attention head. This headwise design fragments recall I/O into transfers as small as one KV block for one head in one layer, only 8 KB under typical settings, severely degrading PCIe bandwidth utilization. It also requires every KV head to keep its own digest in GPU memory; with small block sizes, these per-head digests become a noticeable fraction of the GPU memory footprint. This raises a systems question: is per-head selection truly necessary for accurate recall, or can it be simplified for better I/O efficiency?

Prior work shows that attention heads in long-context LLMs naturally specialize into two roles~\cite{duoattention,razorattention}. \textbf{Streaming heads} mainly attend to sink tokens and local tokens, which we always keep GPU‑resident and do not require recall. \textbf{Retrieval heads} attend to broad middle‑context tokens, which contain semantically rich information and may reside in CPU memory. Since KV offloading only needs to select recall‑relevant middle tokens, sparse selection should be driven primarily by retrieval heads.
Moreover, IMPRESS~\cite{impress} finds that token-importance rankings are highly similar across query heads, with Jaccard overlap between retrieval heads often exceeding 0.8 on OPT-family models.
These observations suggest that per-head independent selection may be unnecessary: a well-chosen retrieval head could guide recall for all KV heads in the layer.

To identify such heads, we define \textbf{Middle Attention Mass (MAM)}. For each query head $q$, we compute its attention mass over middle tokens:
\begin{equation}
\label{eq:mam}
\text{MAM}_q = \sum_{i \in \text{middle}} A_q(i),
\end{equation}
where $A_q(i)$ is the attention weight assigned to middle position $i$. For each KV head $h$, its MAM is the average $\text{MAM}_q$ over all query heads mapped to $h$ by GQA (Grouped-Query Attention)~\cite{gqa}. A high-MAM KV head is a strong retrieval head because its associated query heads focus on middle tokens that may require recall; a low-MAM KV head behaves more like a streaming head and contributes little to recall selection.

Figure~\ref{fig:mam-heatmap} plots the per-head MAM across all layers of Qwen-3 14B, measured on two different datasets: Qasper~\cite{qasper}, a question-answering dataset, and QMSum~\cite{qmsum}, a summarization dataset.
Two properties stand out. First, MAM varies substantially across heads within the same layer: the gap between the highest- and lowest-MAM heads ranges from 1.5× to 12×, averaging 3.5× across layers. This confirms that head specialization is a structural property of the model.
Second, the MAM patterns are stable across the two evaluated datasets, suggesting that head roles are largely model-structural rather than input-specific. Therefore, representative heads can be selected offline once per model.

Building on this insight, we design \textbf{SoloHead sparse selection}, which retains block digests for only one strong retrieval KV head per layer. During decoding, this representative head selects the top-$k$ blocks, and the selected block set is shared by all KV heads in the layer. By replacing per-head selection with a single layer-level decision, SoloHead coalesces recall I/O from head-wise transfers into token-wise transfers, eliminates per-head digest storage, and reduces sparse-selection computation while preserving the retrieval behavior needed for sparse attention.

To quantify the accuracy loss introduced by this shared selection, we use a \textbf{logit-difference} metric. For each layer $L$, we apply sparse attention only to that layer while keeping all other layers in full attention, and compare the final output logits with full attention:
\begin{equation}
\label{eq:logit-diff}
D^{L} = KL(p_{\text{full}}, p_{\text{sparse}}^{L}),
\end{equation}
where $p_{\text{full}}$ is the output probability distribution produced by full attention, and $p_{\text{sparse}}^{L}$ is the output distribution when sparse attention is applied only at layer $L$. $KL(\cdot)$ denotes the Kullback-Leibler divergence~\cite{kl-divergence}.
Since this metric is computed on the final output distribution, it directly measures the perturbation caused by sparse selection in layer $L$. A smaller $D^L$ means the sparse result better preserves full-attention behavior.

We evaluate four selection strategies:

\begin{itemize}
\item $D_{\text{headwise}}$: each query head independently selects its own top-$k$ tokens, representing conventional headwise methods such as ArkVale.
\item $D_{\text{average}}$: query-digest scores are averaged across all heads, and the top-$k$ tokens under the averaged score are shared by all heads.
\item $D_{\text{SoloHead}}$: the KV head with the highest MAM selects the top-$k$ tokens, and the selected token set is shared by all heads.
\item $D_{\text{SoloHead-reversed}}$: the same as SoloHead, but using the \emph{lowest}-MAM KV head.
\end{itemize}

Figure~\ref{fig:logit-diff} reports these metrics on Qwen-3 14B across layers. $D_{\text{SoloHead}}$ is close to $D_{\text{headwise}}$ and $D_{\text{average}}$ for most layers, showing that a single strong retrieval head is sufficient to guide shared top-$k$ selection.
In contrast, $D_{\text{SoloHead-reversed}}$ consistently incurs the largest divergence, confirming that MAM effectively distinguishes retrieval-oriented heads from non-retrieval heads.

\subsubsection{Gather-Scatter Transfer Engine}\label{sec:gather-scatter}

\begin{figure}[tb]
    \centering
    \begin{minipage}[t]{0.48\columnwidth}
        \centering
        \includegraphics[width=\textwidth]{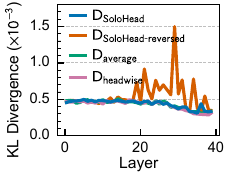}
        \captionof{figure}{Logit-difference across sparse-selection methods.}
        \label{fig:logit-diff}
    \end{minipage}
    \hfill
    \begin{minipage}[t]{0.48\columnwidth}
        \centering
        \includegraphics[width=\textwidth]{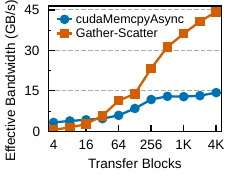}
        \captionof{figure}{Effective bandwidth of \texttt{cudaMemcpyAsync} and gather-scatter transfer.}
        \label{fig:gather-scatter-bw}
    \end{minipage}
\end{figure}

\begin{figure*}[t]
  \centering
  \includegraphics[width=\linewidth]{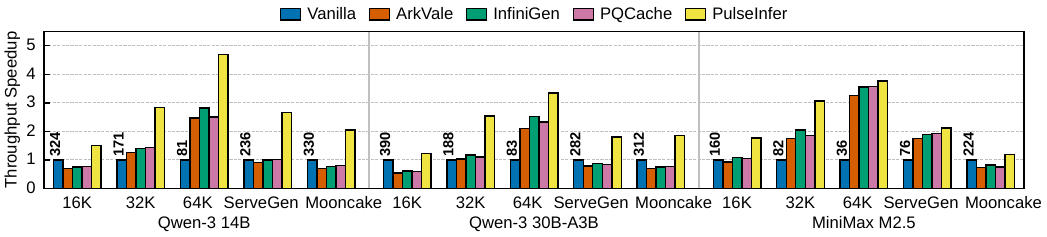}
  \caption{End-to-end decode throughput speedup across models and workloads.}
  \label{fig:eval-e2e-tps}
\end{figure*}

\begin{figure*}[t]
  \centering
  \includegraphics[width=\linewidth]{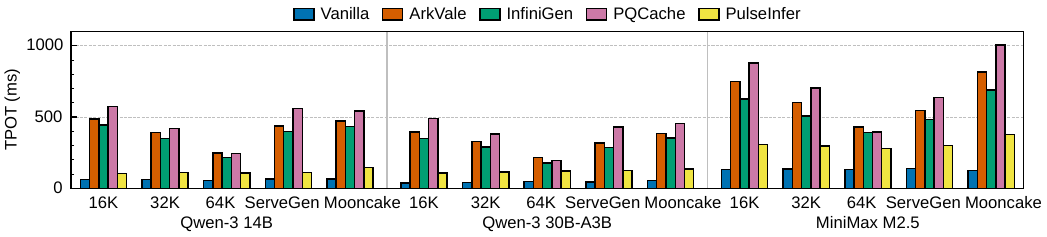}
  \caption{End-to-end TPOT across models and workloads.}
  \label{fig:eval-e2e-tpot}
\end{figure*}

To further improve recall I/O efficiency, PulseInfer employs a gather-scatter transfer engine. Since recalled KV blocks are scattered across DRAM, issuing one DMA per block would create many small PCIe transfers with poor bandwidth utilization. PulseInfer instead gathers the requested blocks into a contiguous pinned CPU staging buffer, transfers them to a GPU staging buffer with one large DMA, and then uses a lightweight GPU scatter kernel to place them into the GPU KV cache. Although this adds extra CPU gather and GPU scatter copies, these copies run at DRAM and HBM bandwidth, respectively, and are much faster than fragmented PCIe transfers. As shown in Figure~\ref{fig:gather-scatter-bw}, when transferring more than 512 blocks, gather-scatter achieves over 2× higher effective bandwidth than threaded \texttt{cudaMemcpyAsync} under a block size of 16.

\subsection{Implementation}\label{sec:impl}

We implement PulseInfer on SGLang~\cite{sglang}, a widely used LLM inference engine. The entire implementation is around \textasciitilde 9 K lines of code.

\textbf{IRQ scheduling.}
PulseInfer implements IRQ scheduling with a two-level scheduler. The outer scheduler partitions requests into a normal batch and an offload batch, controlling which requests are assigned to each batch. It is invoked only when a batch completes a decode step, where it may adjust the batch composition. The inner scheduler operates at layer granularity and uses layer boundaries as cooperative context-switch points. When switching from one batch to another, the inner scheduler records the suspended batch state, including the current layer index, intermediate activations, and forward metadata.

\textbf{Zero-sync I/O engine.}
PulseInfer uses zero-sync I/O submission to remove recall submission from the GPU scheduling critical path (Figure~\ref{fig:gather-scatter}). A small Unified Virtual Memory (UVM)~\cite{uvm} buffer is shared between the GPU and CPU to store a ready flag and the selected KV block indices. After sparse selection, the selection kernel writes the selected indices into this buffer and sets the ready flag. A dedicated CPU polling thread detects the flag, constructs the corresponding recall task, and asynchronously submits it to I/O worker threads. As a result, the main inference thread can continue enqueuing normal-batch kernels without waiting for recall-task construction or submission.

\section{Evaluation}

\textbf{Models.} We evaluate three frontier LLMs: Qwen-3 14B~\cite{qwen3}, Qwen-3 30B-A3B (a Mixture-of-Experts~\cite{moe} model with 30B total parameters and 3B active per token), and MiniMax M2.5 (230B-A10B)~\cite{minimax-m2.5}. These models span total parameter counts from 14B to 230B, cover both dense and MoE architectures, and represent two distinct model families.

\textbf{Setup.} All experiments are conducted on a single GPU server equipped with an Intel Xeon Platinum 8470Q CPU, 1 TB of DRAM, and four NVIDIA RTX PRO 6000 GPUs (96 GB each) connected via PCIe 5.0 ×16. For Qwen‑3 14B and Qwen‑3 30B‑A3B, we use a single GPU; for MiniMax‑M2.5, we use tensor parallelism across all four GPUs.

\textbf{Workloads.} We evaluate performance on two types of workloads: (a) synthetic workloads generated from truncated log-normal distributions with a pronounced right skew in input length. Such distributions closely match the characteristics of realistic traces described in ServeGen~\cite{servegen}. We generate multiple synthetic workloads under different mean input lengths; (b) realistic workloads that replay open-source industrial traces from Mooncake~\cite{mooncake} and ServeGen~\cite{servegen}, each containing the timestamp, input length, and output length of every request.

\textbf{Baselines.} We consider a diverse set of baselines:
\begin{itemize}[leftmargin=*,topsep=0pt]
  \item \textbf{Vanilla}. Standard SGLang inference with full attention and GPU-resident KV cache.
  \item \textbf{ArkVale~\cite{arkvale}}. Sparse attention with KV cache offloading; identifies important KV entries using lightweight attention metadata and performs synchronous recall I/O.
  \item \textbf{InfiniGen~\cite{infinigen}}. Sparse attention with KV cache offloading; predicts important KV entries ahead of execution and prefetches recall I/O one layer ahead.
  \item \textbf{PQCache~\cite{pqcache}}. Sparse attention with KV cache offloading; uses product quantization~\cite{pq-quant} and asymmetric distance computation (ADC) to identify important KV entries.
\end{itemize}

We implement ArkVale, InfiniGen, and PQCache in SGLang to the best of our ability.
Since our evaluation focuses on the decoding stage, we skip the prefill phase to emulate a decode instance in a prefill-decode disaggregated setting.

Unless otherwise specified, each offloaded request is assigned a 2,048-token GPU KV-cache budget, comprising 16 sink tokens, 128 local tokens, and 1,904 important middle tokens. Sink and local tokens remain resident on the GPU, while only the selected middle tokens are subject to CPU-to-GPU recall. The block size is set to 16, and the CPU-side KV cache capacity is set to 10× the GPU-side capacity.

\subsection{End-to-End Performance}

\subsubsection{Overall Throughput}

We first evaluate end-to-end decode throughput under both synthetic and realistic workloads. For the synthetic workloads, the mean input lengths are set to 16K, 32K, and 64K tokens. Figure~\ref{fig:eval-e2e-tps} reports throughput speedup over Vanilla across three models.

PulseInfer consistently achieves the highest throughput across all models and workloads, improving decode throughput by 1.5x--4.7× over Vanilla and 1.1×--2.6× over the best existing offloading baseline. The gains increase with input length because Vanilla becomes increasingly constrained by GPU KV-cache capacity. For example, on Qwen-3 14B, when the mean input length grows from 16K to 64K, Vanilla throughput drops from 324 tok/s to 81 tok/s, while PulseInfer only decreases from 488 tok/s to 380 tok/s. This is because IOAA admits more offloaded requests under tighter GPU memory pressure, while IRQ hides the additional recall I/O behind normal-batch computation.

Existing offloading baselines benefit from larger CPU-side KV capacity, but their throughput remains limited by blocking and inefficient recall I/O. On shorter workloads, this overhead can even outweigh the batching benefit; for instance, at 16K mean input length on Qwen-3 14B, the best offloading baseline reaches only 0.7× the throughput of Vanilla. InfiniGen generally performs best among prior offloading systems because one-layer-ahead prefetching hides part of the recall latency, but its fixed overlap window cannot robustly cover recall volume that varies across layers, requests, and batches.

Across models, PulseInfer obtains the largest relative gain on Qwen-3 14B. Compared with the MoE models, Qwen-3 14B has more activated computation per token, giving IRQ more layer computation to overlap each recall transfer and reducing the required overlap depth. The relative gain is smaller on MiniMax M2.5 because it is served with tensor parallelism across four GPUs, which distributes KV heads and effectively increases aggregate recall bandwidth for all offloading methods. Nevertheless, PulseInfer remains the fastest across all evaluated models.

PulseInfer also maintains strong gains on realistic ServeGen and Mooncake traces, where random arrivals and time-varying input-length distributions continuously change the normal/offload mix and aggregate recall-I/O load. Static offloading policies and fixed prefetch pipelines become unstable under such drift.
In contrast, PulseInfer combines IRQ to hide variable recall latency at layer granularity with IOAA to dynamically adjust the normal/offload split under changing workload and memory pressure.

\begin{figure}[tb]
    \centering
    \begin{minipage}[t]{0.48\columnwidth}
        \centering
        \includegraphics[width=\textwidth]{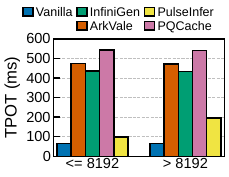}
        \captionof{figure}{TPOT breakdown by request length.}
        \label{fig:eval-short-long-tpot}
    \end{minipage}
    \hfill
    \begin{minipage}[t]{0.48\columnwidth}
        \centering
        \includegraphics[width=\linewidth]{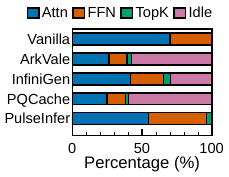}
        \captionof{figure}{Latency breakdown.}
        \label{fig:latency-breakdown}
    \end{minipage}
\end{figure}

\subsubsection{Per-Token Latency}

\begin{figure}[tb]
    \centering
    \begin{minipage}[t]{0.48\columnwidth}
        \centering
        \includegraphics[width=\textwidth]{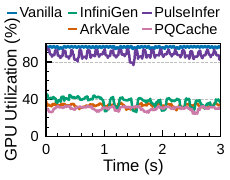}
        \captionof{figure}{GPU utilization across different methods.}
        \label{fig:all-gpu-util}
    \end{minipage}
    \hfill
    \begin{minipage}[t]{0.48\columnwidth}
        \centering
        \includegraphics[width=\linewidth]{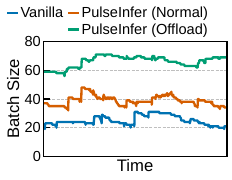}
        \captionof{figure}{Batch size during decoding.}
        \label{fig:eval-batch-size}
    \end{minipage}
\end{figure}

Figure~\ref{fig:eval-e2e-tpot} reports the mean TPOT in the end-to-end evaluation. Across all models and workloads, InfiniGen is the strongest existing offloading baseline, while PulseInfer consistently achieves lower TPOT than all offloading baselines. Compared with InfiniGen, PulseInfer reduces TPOT by 16.0\%--76.4\%, corresponding to 1.19--4.23× lower per-token latency. This confirms that PulseInfer improves end-to-end throughput not by blindly increasing the number of offloaded requests, but by substantially reducing the per-token stalls introduced by recall I/O.

Compared with Vanilla, PulseInfer increases mean TPOT by 2.2× on average. This is expected because Vanilla keeps all KV cache on the GPU and thus incurs neither recall I/O nor normal/offload batch switching overhead. PulseInfer therefore intentionally trades a moderate increase in per-token latency for substantially higher throughput. We consider this tradeoff acceptable for long-context decode serving: Vanilla achieves low TPOT only by admitting very few concurrent requests under GPU memory pressure, whereas PulseInfer improves decode throughput by up to 4.7× while keeping TPOT much lower than existing offloading systems.

Additionally, Figure~\ref{fig:eval-short-long-tpot} breaks down TPOT by request length on Qwen-3 14B under the Mooncake trace. For requests shorter than 8K tokens, PulseInfer increases TPOT by only 1.5× over Vanilla; for requests longer than 8K tokens, the increase is 2.3×. In contrast, existing offloading baselines increase TPOT by 6.7--8.3× for both short and long requests. This demonstrates that PulseInfer effectively isolates short requests from recall-I/O stalls through IRQ-style scheduling.

\subsection{Performance Analysis}

In this section, we conduct a series of experiments and measurements to analyze the performance of PulseInfer. Unless otherwise specified, all evaluations in this section are performed on Qwen-3 14B using the ServeGen trace.

\begin{figure}[tb]
    \centering
    \begin{minipage}[t]{0.48\columnwidth}
        \centering
        \includegraphics[width=\textwidth]{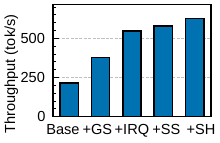}
        \captionof{figure}{Ablation study. Base denotes ArkVale, a synchronous offloading baseline.}
        \label{fig:eval-ablation}
    \end{minipage}
    \hfill
    \begin{minipage}[t]{0.49\columnwidth}
        \centering
        \includegraphics[width=\linewidth]{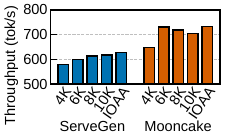}
        \captionof{figure}{Throughput comparison of IOAA and fixed admission thresholds.}
        \label{fig:eval-ioaa}
    \end{minipage}
\end{figure}

\begin{figure}[tb]
    \centering
    \begin{minipage}[t]{0.48\columnwidth}
        \centering
        \includegraphics[width=\textwidth]{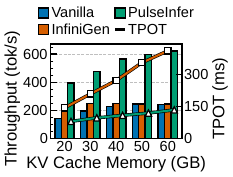}
        \captionof{figure}{Impact of GPU KV cache capacity on throughput and TPOT.}
        \label{fig:eval-memory-capacity-impact}
    \end{minipage}
    \hfill
    \begin{minipage}[t]{0.48\columnwidth}
        \centering
        \includegraphics[width=\linewidth]{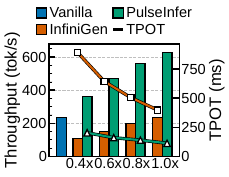}
        \captionof{figure}{Impact of PCIe bandwidth (relative to PCIe 5x16).}
        \label{fig:eval-pcie-bw-impact}
    \end{minipage}
\end{figure}

\begin{table*}[!t]
\centering
\footnotesize
\setlength{\tabcolsep}{3pt}
\caption{Accuracy comparison across Document QA, Summarization, Few-shot Learning, and Retrieval benchmarks.}
\label{tab:accuracy}
\begin{tabular}{llcccccccccc}
\toprule
\multirow{2}{*}[-2pt]{\textbf{Model}} & \multirow{2}{*}[-2pt]{\textbf{Method}} & \multicolumn{3}{c}{\textbf{Document QA}} & \multicolumn{2}{c}{\textbf{Summarization}} & \multicolumn{2}{c}{\textbf{Few-shot Learning}} & \multicolumn{2}{c}{\textbf{Retrieval}} & \multirow{2}{*}[-2pt]{\textbf{\shortstack{Avg.\\Score}}} \\
\cmidrule(lr){3-5} \cmidrule(lr){6-7} \cmidrule(lr){8-9} \cmidrule(lr){10-11}
& & Qasper & 2WikiMQA & HotpotQA & GovReport & VCSUM & TREC & TriviaQA & RetrievalZH & RetrievalEN & \\
\midrule
\multirow{6}{*}[0pt]{\textbf{Qwen-3 14B}}
& Full Attention & \textbf{47.97} & 74.45 & 67.38 & 29.16 & 20.80 & 75.63 & 90.82 & \textbf{100.00} & \textbf{100.00} & 67.36 \\ \cline{2-12}
& ArkVale & 46.13 & \textbf{76.61} & \textbf{70.57} & 29.39 & \textbf{21.00} & 74.11 & 90.61 & 99.50 & \textbf{100.00} & 67.55 \\ \cline{2-12}
& InfiniGen & 44.81 & 75.22 & 69.39 & \textbf{29.50} & 20.58 & 77.15 & 90.22 & 99.50 & 99.50 & 67.32 \\ \cline{2-12}
& PQCache & 46.84 & 75.78 & 70.29 & 29.05 & 20.88 & 76.14 & \textbf{91.38} & 99.50 & 99.50 & 67.71 \\ \cline{2-12}
& \textbf{PulseInfer} & 46.88 & 76.06 & 69.78 & 29.28 & 20.64 & \textbf{77.16} & 90.45 & \textbf{100.00} & 99.50 & \textbf{67.75} \\ \cline{2-12}
&   \quad +reversed rank & 43.75 & 72.80 & 64.67 & 28.34 & 19.72 & 71.57 & 90.40 & 99.50 & 97.50 & 65.36 \\
\midrule
\multirow{6}{*}[0pt]{\textbf{Qwen-3 30B-A3B}}
& Full Attention & 48.12 & 77.54 & 66.89 & 28.73 & 20.71 & 77.16 & 92.17 & \textbf{100.00} & 81.50 & 65.87 \\ \cline{2-12}
& ArkVale & 47.22 & 75.23 & 67.05 & 29.05 & \textbf{21.47} & \textbf{80.20} & 91.92 & 99.00 & \textbf{91.50} & 66.96 \\ \cline{2-12}
& InfiniGen & 48.49 & 78.51 & 66.82 & 28.81 & 21.10 & 79.19 & 91.33 & 99.50 & 82.00 & 66.19 \\ \cline{2-12}
& PQCache & 47.65 & \textbf{78.75} & 66.55 & 29.10 & 21.19 & 78.68 & 92.00 & 99.50 & 90.50 & 67.10 \\ \cline{2-12}
& \textbf{PulseInfer} & \textbf{48.73} & 78.03 & \textbf{68.60} & \textbf{29.12} & 21.12 & 76.65 & \textbf{92.70} & \textbf{100.00} & \textbf{91.50} & \textbf{67.38} \\ \cline{2-12}
&   \quad +reversed rank & 46.27 & 76.22 & 63.51 & 28.43 & 20.02 & 71.57 & 91.17 & 99.50 & 83.50 & 64.47 \\
\midrule
\multirow{6}{*}[0pt]{\textbf{\shortstack{MiniMax M2.5\\(230B-A10B)}}}
& Full Attention & 50.10 & \textbf{80.57} & \textbf{72.61} & 33.04 & 23.94 & 78.43 & 91.67 & \textbf{100.00} & \textbf{100.00} & \textbf{70.04} \\ \cline{2-12}
& ArkVale & \textbf{51.20} & 79.88 & 71.89 & 32.80 & 23.94 & 76.65 & 91.61 & \textbf{100.00} & \textbf{100.00} & 69.77 \\ \cline{2-12}
& InfiniGen & 49.30 & 79.80 & 70.61 & \textbf{33.06} & \textbf{24.32} & \textbf{80.20} & 91.83 & \textbf{100.00} & \textbf{100.00} & 69.90 \\ \cline{2-12}
& PQCache & 50.17 & 78.66 & 70.15 & 33.05 & 23.94 & 77.16 & 91.92 & \textbf{100.00} & \textbf{100.00} & 69.45 \\ \cline{2-12}
& \textbf{PulseInfer} & 51.07 & 79.93 & 71.99 & 33.02 & 23.96 & 77.66 & \textbf{91.98} & \textbf{100.00} & \textbf{100.00} & 69.96 \\ \cline{2-12}
&   \quad +reversed rank & 50.10 & 76.83 & 64.77 & 30.91 & 22.15 & 71.83 & 91.82 & 98.50 & 95.75 & 66.96 \\
\bottomrule
\end{tabular}
\end{table*}

\textbf{GPU utilization and latency breakdown.}
We break down decoding latency and measure runtime GPU utilization, as shown in Figures~\ref{fig:latency-breakdown} and~\ref{fig:all-gpu-util}. ArkVale and PQCache synchronously wait for recall I/O, so recall stalls dominate their latency and keep GPU utilization low.
InfiniGen tries to hide recall latency with one-layer-ahead K/V prefetching, but the fixed prefetch window cannot keep up with dynamic recall I/O and still leaves noticeable idle periods. In contrast, PulseInfer uses IRQ scheduling to overlap recall I/O with normal GPU execution, largely eliminating recall-induced idle time and sustaining GPU utilization close to Vanilla.

\textbf{Batch size.}
We record the normal- and offload-batch sizes during inference. As shown in Figure~\ref{fig:eval-batch-size}, PulseInfer achieves a total batch size 4.7× larger than Vanilla, with both batches individually exceeding the Vanilla batch size. This is because Vanilla must keep all KV caches on the GPU, so a few long requests can quickly exhaust GPU memory. PulseInfer offloads long requests to CPU memory, releasing GPU capacity for more short requests in the normal batch while further expanding serving capacity with the offload batch.

\textbf{Ablation study.}
Figure~\ref{fig:eval-ablation} shows the ablation study of PulseInfer, where GS denotes gather-scatter transfer (\S\ref{sec:gather-scatter}), IRQ denotes interruptible layer-wise scheduling (\S\ref{sec:irq}), SS denotes semi-synchronous interrupt detection (\S\ref{sec:semi-sync}), and SH denotes SoloHead (\S\ref{sec:solohead}). All components contribute to end-to-end throughput improvement. Together, they raise throughput to over 3$\times$ that of Vanilla, demonstrating that PulseInfer's gains come from the combined effect of more efficient recall transfers, adaptive I/O hiding, lower scheduling overhead, and coalesced sparse selection.

\textbf{Effectiveness of IOAA.}
We compare IOAA with fixed short/long admission thresholds on the ServeGen and Mooncake traces. As shown in Figure~\ref{fig:eval-ioaa}, IOAA consistently achieves the highest throughput on both traces. Moreover, the best fixed threshold differs between the two traces, indicating that a static admission policy cannot generalize across workload distributions. In contrast, IOAA dynamically searches for the best offloading admission decision at runtime, allowing PulseInfer to adapt to workload-dependent memory pressure and recall-I/O demand.

\textbf{Impact of GPU memory size.}
Figure~\ref{fig:eval-memory-capacity-impact} shows throughput and TPOT under different available GPU KV cache capacities. For InfiniGen, larger capacity does not improve throughput: it admits more offloaded requests, but the resulting recall I/O increases GPU idle time, as reflected by its nearly linear TPOT growth. This indicates that InfiniGen is bottlenecked by recall I/O, so larger batches cannot translate into higher throughput. In contrast, PulseInfer continues to improve throughput with higher KV cache capacity because IRQ scheduling hides recall latency and prevents the system from becoming I/O-bound.

\textbf{Impact of PCIe bandwidth.}
We throttle the available I/O bandwidth to quantify the impact of PCIe bandwidth. Figure~\ref{fig:eval-pcie-bw-impact} reports throughput and TPOT under different bandwidth scaling factors. Higher bandwidth improves both InfiniGen and PulseInfer by shortening recall I/O. However, PulseInfer remains faster than Vanilla even at 0.4× bandwidth, whereas InfiniGen only matches Vanilla under unconstrained bandwidth. Moreover, at 0.4× bandwidth, InfiniGen's TPOT increases by 2.2×, while PulseInfer's increases by only 1.7×. This shows that PulseInfer is more robust to limited I/O bandwidth thanks to IRQ scheduling.

\subsection{Accuracy}

We evaluate the accuracy of PulseInfer on a diverse set of LLM benchmarks spanning multiple domains. Specifically, we use LongBench~\cite{longbench}, a widely used benchmark suite for long-context LLMs, covering document question answering, meeting and report summarization, few-shot learning, and passage retrieval tasks. Table~\ref{tab:accuracy} reports the results on three models. Each dataset is evaluated using its corresponding metric, and higher scores indicate better accuracy.
We enable chain-of-thought reasoning for all three models and set the thinking budget to 2048 tokens for every dataset.

As shown in Table~\ref{tab:accuracy}, PulseInfer achieves near-lossless accuracy across benchmarks. On the two Qwen-3 models, PulseInfer even slightly outperforms full attention.
This indicates that a strong retrieval head identifies important middle-context tokens more accurately, avoiding the noisy selections that streaming heads would otherwise introduce. As a result, the shared top‑$k$ blocks capture the recall‑relevant context needed for sparse attention more effectively.

We further validate the MAM-based head selection by evaluating a reversed configuration, where PulseInfer chooses the lowest-MAM KV head as the SoloHead instead of the highest-MAM head. As shown by the reversed rank result in Table~\ref{tab:accuracy}, this choice causes a substantial accuracy drop. This confirms that MAM effectively separates retrieval-oriented heads from streaming heads, and that middle-context recall should be guided by retrieval heads rather than heads that primarily attend to GPU-resident sink and local tokens.

\section{Related Work}

\textbf{KV cache offloading.}
ArkVale~\cite{arkvale} backs up evicted KV pages in CPU memory and recalls them using compact page digests. InfiniGen~\cite{infinigen} predicts next-layer important KV entries and prefetches them ahead of use. PQCache~\cite{pqcache} formulates KV selection as approximate retrieval and uses product quantization to search offloaded cache. These systems optimize KV selection, indexing, caching, or prefetching, but recall I/O can still stall execution under dynamic demand and fragmented transfers. PulseInfer addresses this missing systems layer with an I/O-centric design that jointly hides, controls, and coalesces sparse KV recalls.

\textbf{Fine-grained scheduling for LLM serving.}
Sarathi-Serve~\cite{sarathi-serve} chunks long prefills and interleaves them with decodes, NanoFlow~\cite{nanoflow} uses operation-level nano-batches to overlap compute, memory, and communication, and NEO~\cite{neo} overlaps CPU-side decoding attention with GPU execution through asymmetric CPU--GPU pipelining. These systems refine scheduling granularity for batching efficiency, prefill--decode overlap, or CPU--GPU load balance. PulseInfer instead schedules at layer boundaries to react to sparse-recall I/O completion and hide unpredictable PCIe latency while keeping attention computation on the GPU.

\textbf{Head-aware sparse attention.}
MInference~\cite{minference} exploits head-specific sparse patterns, DuoAttention~\cite{duoattention} separates retrieval heads from streaming heads, and LServe~\cite{lserve} combines head-aware sparsity with query-centric KV-page selection. While these works use head specialization to reduce attention cost, PulseInfer uses it to coalesce KV-cache recall I/O: SoloHead selects one retrieval-oriented head per layer to guide block selection for all KV heads.

\section{Conclusion}

We present PulseInfer, an I/O-centric sparse KV cache offloading system for long-context LLM decoding. PulseInfer combines interruptible layer-wise scheduling, IO-Adaptive Offloading Admission and SoloHead sparse selection to explicitly schedule, control, and coalesce recall I/O. Evaluation shows that PulseInfer improves decode throughput and GPU utilization while preserving near-lossless accuracy.

\bibliographystyle{ACM-Reference-Format}
\bibliography{reference}

\end{document}